\documentclass{article}

\usepackage[final]{nips_2017}
\usepackage{graphicx}
\usepackage[utf8]{inputenc} 
\usepackage[T1]{fontenc}    
\usepackage{hyperref}       
\usepackage{url}            
\usepackage{booktabs}       
\usepackage{amsfonts}       
\usepackage{nicefrac}       
\usepackage{microtype}      

\title{Visual Transfer between Atari Games using Competitive Reinforcement Learning }

\author{
  Akshita Mittel \\
  \And
  Purna Sowmya Munukutla \\
  \And
  Himanshi Yadav \\
  \AND 
  \texttt{\{amittel, spmunuku, hyadav\}@andrew.cmu.edu} \\
  Robotics Institute \\
  Carnegie Mellon University
}

\begin{document}

\maketitle

\begin{abstract}
This paper explores the use of deep reinforcement learning agents to transfer knowledge from one environment to another. More specifically, the method takes advantage of asynchronous advantage actor critic (A3C) architecture to generalize a target game using an agent trained on a source game in Atari. Instead of fine-tuning a pre-trained model for the target game, we propose a learning approach to update the model using multiple agents trained in parallel with different representations of the target game. Visual mapping between video sequences of transfer pairs is used to derive new representations of the target game; training on these visual representations of the target game improves model updates in terms of performance, data efficiency and stability. In order to demonstrate the functionality of the architecture, Atari games \texttt{Pong-v0} and \texttt{Breakout-v0} are being used from the OpenAI gym environment; as the source and target environment. 
\end{abstract}

\section{Introduction}
Our paper is motivated by 'Learning to Learn' as was discussed in the NIPS-95 workshop, where interest was shown in applying previously gained knowledge to a new domain for learning new tasks \citep{5288526}. The learner uses a series of tasks learned in the source domain, to improve and learn the target tasks. In our paper, a frame taken from the target game is mapped to the analogous state in the source game and the trained policy learned from the source game knowledge is used to play the target game. 

We also rely on transferring knowledge using transfer learning methods that have shown to improve performance and stability \cite{stanford_dudes}. Expert performances can be achieved on several games and with the model
complexity of a single expert. Significant improvements in learning speeds in target games are also seen \cite{stanford_dudes,dudes}.  The same learned weights can be generalized from a source game to several new target games. This treatment of weights from a previously trained model has implications in Safe
Reinforcement Learning \cite{stanford_dudes} as we can easily learn from already learned stable agents.

We find underlying similarities between the source and the target domains i.e.\ different Atari Games to represent common knowledge using  Unsupervised Image-to-image Translation (UNIT) Generative adversarial networks (GANs) \cite{GAN}. 

Recent progress in the applications of Reinforcement Learning (RL) using deep networks has led us to policy gradient methods like the A3C algorithm, that can autonomously achieve human-like performance in Atari games\cite{DBLP:journals/corr/MnihBMGLHSK16}. In an A3C network, several agents are executed in parallel, with different starting policies and the global state updated intermittently by the agents. A3C has a smaller computation cost and we do not have to calculate the Q value for every action in the action space to find the maximum. 

\citeauthor{competitive} propose a training method where two games are simultaneously trained where each game within the neural network competes for representational space. In our paper, the target game competes with its visual representation obtained after using the UNIT GAN as a visual mapper between the source and target game.

\section{Related Work}
Translating images to another set of images has been extensively studied by the vision and the 
graphics communities. Recently, considerable interest has been shown in the field of unsupervised image translation \cite{DBLP:journals/corr/ZhuPIE17, DBLP:journals/corr/KimCKLK17} which successfully aids in the task of domain adaptation.  The main objective is to learn a mapper that translates an input image to an output image without prior aligned image pairs. By assuming that the input and the output images have some underlying relationship, in \cite{DBLP:journals/corr/0001T16} introduce CoGAN and cross-modal scene networks \cite{cross-modal}  that learn a common representation using a weight sharing technique. \citeauthor{DBLP:journals/corr/ZhuPIE17} achieve the same without relying on a similarity function between the input and output images and also, not assuming that the input and output images are in the same low-dimensional embedding space.

Domain adaption aims to generate a shared representation for distinct domains \cite{Ganin:2015:UDA:3045118.3045244}. \cite{DBLP:journals/corr/ParisottoBS15} use multi-task and transfer learning to act in a new domain by using previous knowledge. The multi-task learning method employed, called "Actor-Mimic", used model compression techniques to train one multi-task network from several expert networks. The multi-task network is treated as a Deep Q network (DQN) pre-trained on some tasks in the source domain. The DQN which used this multi-task pre-training learns a target task significantly faster than a DQN starting from a random
initialization, effectively demonstrating that the source task representations generalize to the target
task. \cite{competitive} adapt the DQN algorithm to competitively learn two similar Atari games. The authors prove that a DQN trained to solve a particular specific task does not perform well on unforeseen tasks, however, their competitively learning technique does when a DQN agent is trained simultaneously.

More recently, \citeauthor{s.2018domain} use a 2D GridWorld environment which has been rendered differently to create several versions of the same environment that still have underlying similarities and employ multi-task learning and domain adaption on these different versions. The environment is then learned using the A3C algorithm.

\section{Methods}
\label{gen_inst}
A3C is a lightweight asynchronous variant of the actor critic model that shows success in learning a variety of Atari games. A3C network architecture consists of four convolutional layers followed by an LSTM layer and two fully connected layers to predict actions and value functions of the states. However, the entire representational space of the network is specialized for a particular game that it has been trained on. The idea of transfer learning methods is to use this experience to improve the performance of other similar tasks.

The goal of this paper is to use an RL agent to generalize between two related but vastly different Atari games like \texttt{Pong-v0} and \texttt{Breakout-v0}. This is done by learning visual mappers across games: given a frame from the source game, we should be able to generate the analogous frame in the target game. Building on the existence of these mappers, the training method we propose is to simultaneously learn two representations of the target game and effectively making them compete for representational space within the neural network. \\
\subsection{Visual mappers across Atari games}
\begin{figure}[h]
\includegraphics[width= \textwidth]{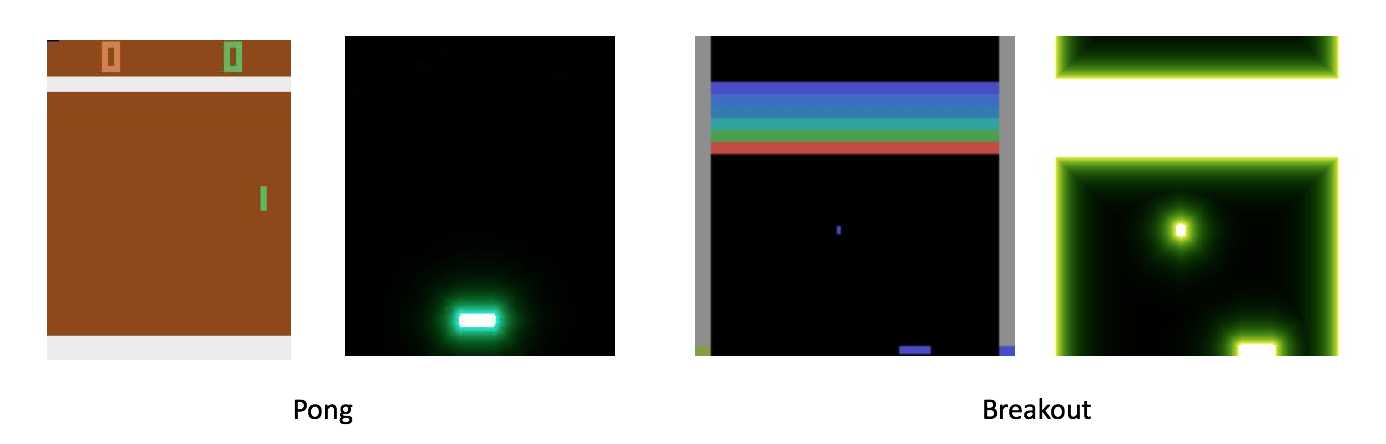}
\caption{Images with input frame and preprocessed frame for source (\texttt{Pong-v0}) and target (\texttt{Breakout-v0}) game}
\label{figure1}
\end{figure}

To create visual analogies between games, we rely on the core ideas explored in \cite{dudes} to learn the mapping $G$ : $s$ $\to$ $t$ between source game ($s$) and target game ($t$) in an unsupervised manner. The unsupervised learning step as described in the setup of \cite{dudes} requires preprocessing of data. The attention maps of input frames are used as the preprocessed frames. Attention maps are generated by rotation of input image so that the main axis of motion is horizontal, binarizing the input after subtracting the median pixel and applying dilation operator on that frame to enlarge relevant object sizes. In the final preprocessing step, output is obtained by cloning dilated image and applying two levels of blurring to create three channels of the image as shown in Figure \ref{figure1}.

To train the mapper function $G$, shared latent space assumption can be made for Atari games and the mapping is trained with an unsupervised image-to-image translation framework. The pre-processed frames from source game and target game are mapped to the same latent representation in a shared latent space across both games. We assume that a pair of corresponding images in two different domains of source game and target game can be inferred by learning two encoding functions, to map images to latent codes and two generation functions that map latent codes to images. Based on this assumption of shared latent space, we use an existing framework that is based on GANs and VAEs to learn the mapper.

\begin{figure}[h]
\includegraphics[width= \textwidth]{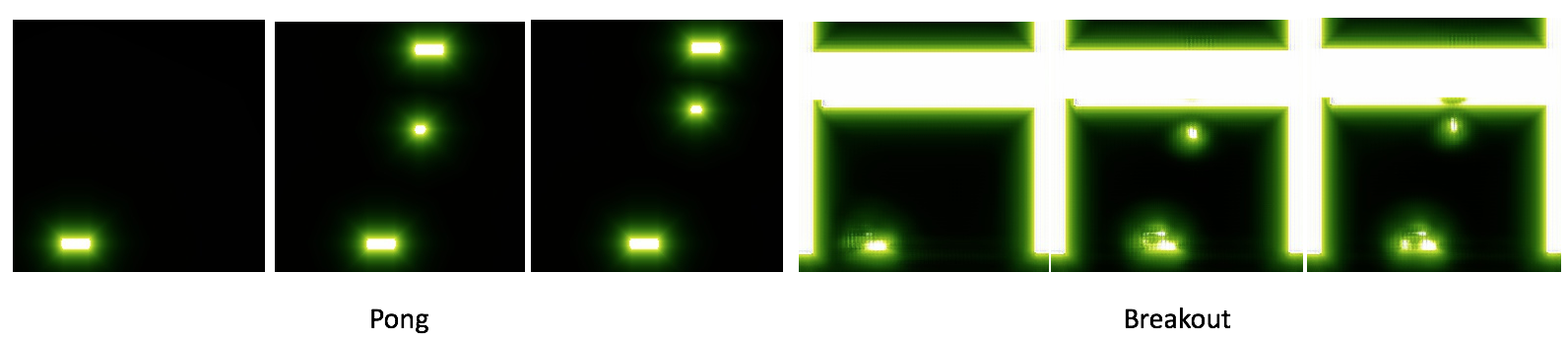}
\caption{Mapped images with preprocessed frames for source (\texttt{Pong-v0}) on the left and target (\texttt{Breakout-v0}) game mapping outputted by trained UNIT GAN}
\label{figure2}
\end{figure}
\begin{figure}[h]
\includegraphics[width= \textwidth]{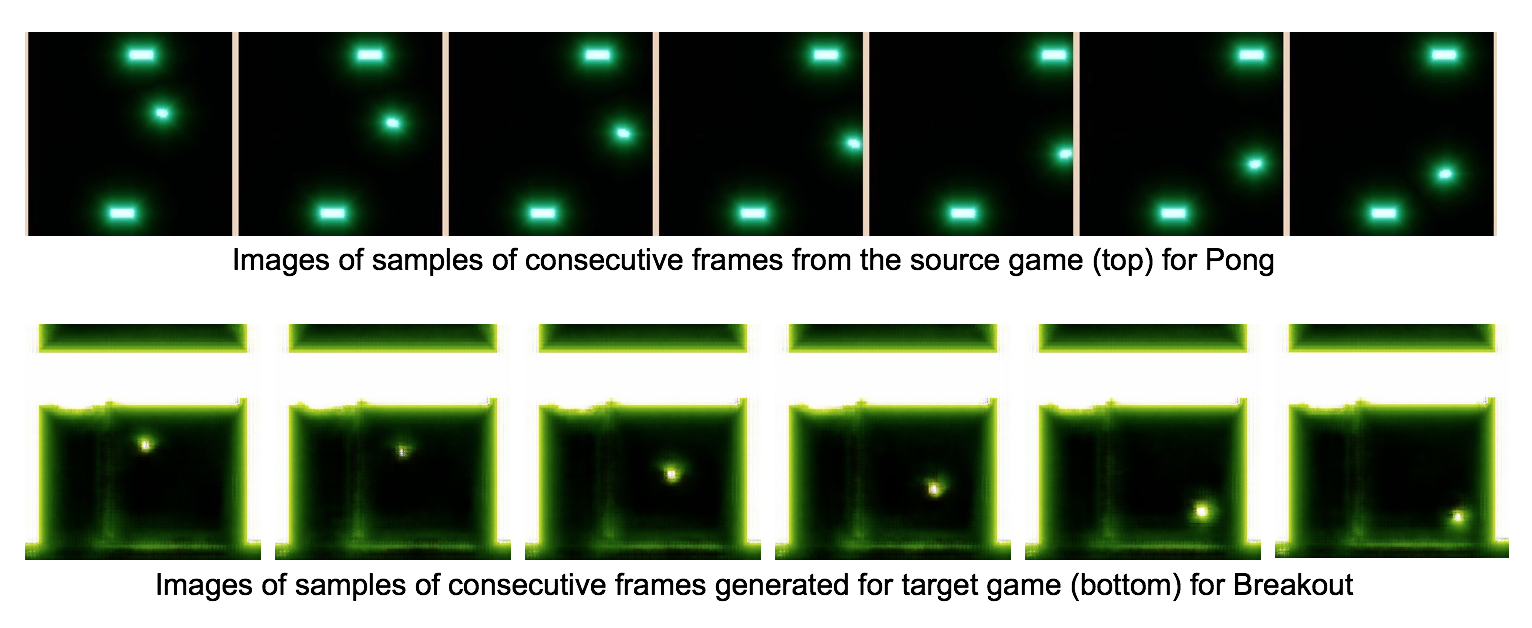}
\caption{Mapped images with preprocessed frames for source (\texttt{Pong-v0}) and target (\texttt{Breakout-v0}) game mapping outputted by trained UNIT GAN as implemented in \cite{dudes}}
\label{figure3}
\end{figure}
The model is implemented with the network architecture of UNIT GAN \citep{GAN} with unit consistency loss. The encoding and generating functions are implemented using CNNs and the shared-latent space assumption is enforced with a weight sharing constraint across these functions. In addition, adversarial discriminators for the respective domains are trained to evaluate whether the translated images are realistic. To generate an image in target domain of \texttt{Breakout-v0}, we use the encoding function of source game to get the latent code and use the generating function of target game to get the frame. The resulting images are shown in Figure \ref{figure1} with the preprocessing step of input frames and a naive mapping between source and target games that has been learned using UNIT GAN. 

\subsection{Transfer Learning Method}
One of the challenges that we are trying to address in this paper is to prove or disprove that visual analogies across games are necessary and sufficient to transfer the knowledge of playing one particular game to another.

In the recent times, policy gradient methods like A3C are shown to be extremely effective in learning the world of Atari games. They are currently set as the baseline for \texttt{Pong-v0} and \texttt{Breakout-v0} games in terms of training time vs rewards and they have already surpassed the performance achieved by Dueling DQN and Double DQN networks for these games. The idea of A3C networks is to use multiple workers in parallel that can each interact with the environment and update the shared model simultaneously. Thus, A3C networks asynchronously exchange multiple agents in parallel instead of using experience replay. 

We use the baseline A3C network trained for source game (\texttt{Pong-v0}) in the first stage of our training process and transfer the knowledge from this model to learning to play target game (\texttt{Breakout-v0}). We measure the efficiency of transfer learning method in terms of training time and data efficiency across parallel actor-learners. In the second stage of training process, we use two representations of the target game amongst the workers in parallel. The first representation of transfer process uses the target game frames taken directly from the environment. The second representation of transfer process uses the frames learned from the visual mapper i.e., $G(S)$. The ratio of number of workers that train directly on frames queried from the target game and frames mapped from the source game is a hyperparameter that is determined through experimentation.

\begin{figure}[h]
\centering
\includegraphics[width= \textwidth]{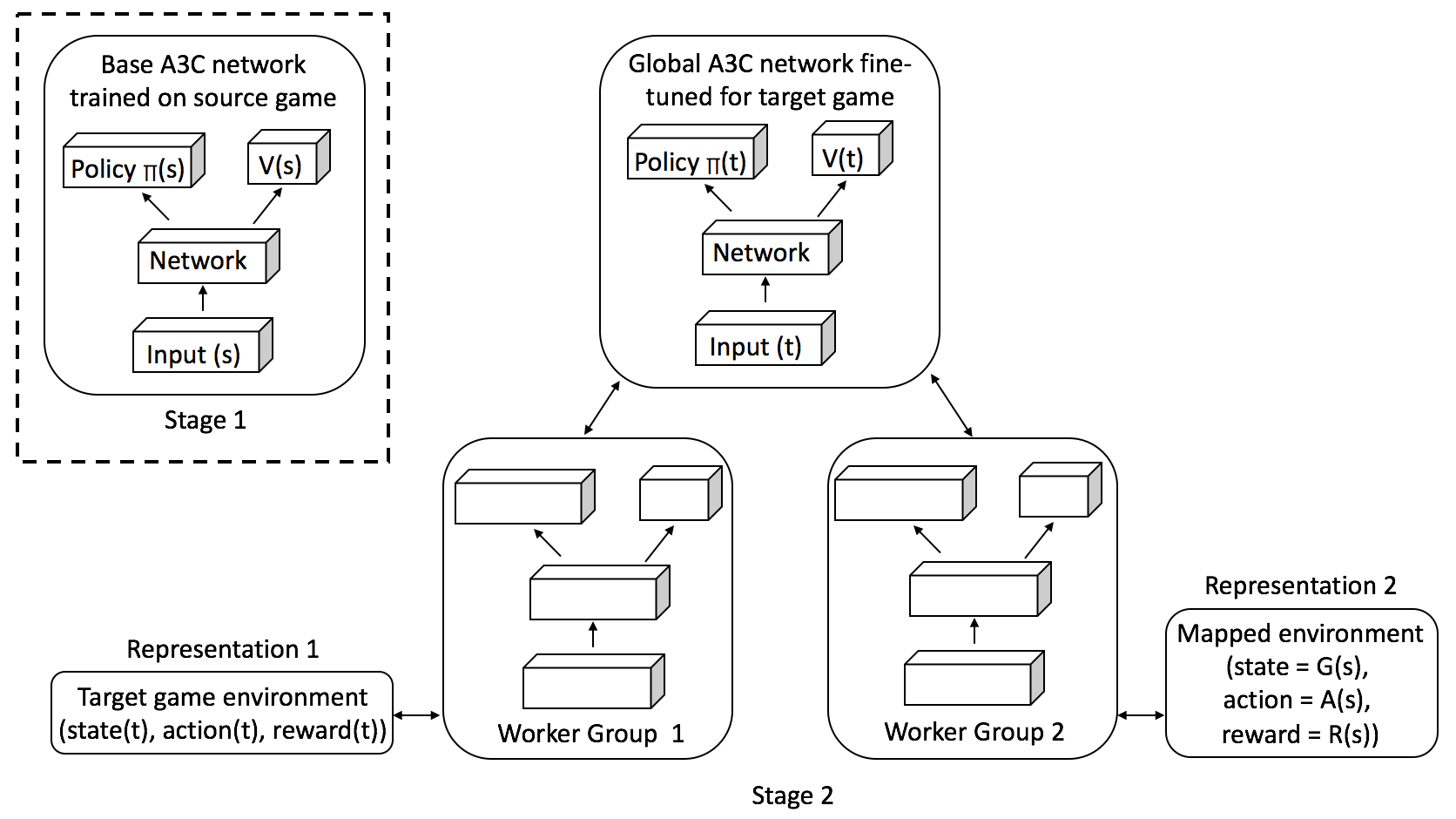}
\caption{Transfer learning process with training for the source game in stage 2 and training for the target game using two different representations in a competitive manner. Representation 1 is for the target game is queried directly from the environment and representation 2 for the target game is extracted using visual transfer of states, static mapping of actions and rewards from source game}
\label{figure4}
\end{figure}

In this way, we transfer the knowledge from source game to target game by competitively and simultaneously fine-tuning the model using two different visual representations of the target game. One visual representation is queried directly from the environment and the other visual representation uses the learned mappers for source game to extract the frames of target game. The actions of the target game for the second representation are determined from a static mapping of actions between source and target games. Since \texttt{Pong-v0} and \texttt{Breakout-v0} have similar game strategies of controlling a paddle to hit the ball to obtain a certain objective, it is intuitive to determine a meaningful static mapping of actions. The six actions of \texttt{Pong-v0} \{No Operation, Fire, Right, Left, Right Fire, Left Fire\} is mapped to four actions actions of \texttt{Breakout-v0} as \{Fire, Fire, Right, Left, Right, Left\} respectively. The rewards are mapped directly from source game to target game without any scaling.  

\section{Results}
\label{headings}
This section describes the results of different stages in our transfer learning pipeline. For this particular pipeline we use the OpenAI gym environment of Atari games and generalize the RL agent between \texttt{Pong-v0} and \texttt{Breakout-v0}.

\subsection{State, Action, and Reward Mappers}

As described in Section \ref{gen_inst}, the first stage is to obtain the preprocessed images for the frames of both the source (\texttt{Pong-v0}) and target (\texttt{Breakout-v0}) games. The next stage is to obtain the state, action, and reward mappings to train our A3C network. In order to get the state mapping $G$ from our source to target frames\cite{dudes} propose training UNIT GAN network architecture. However, this stage requires immense amounts of manual processing steps; generating enough unique state spaces for both the source and target games. Furthermore, it took us approximately six hours of training time to run this architecture on a GeForce GTX 1080 GPU for an epoch. To avoid this bottle-neck, we are currently using pre-trained models that contain the visual mapping from \texttt{Pong-v0} to \texttt{Breakout-v0} provided by the original authors\cite{dudes}. The results of the mapping from the source frames $s$ to $G(s)$ are shown in Figure \ref{figure2} and Figure \ref{figure3}. 

The process of reward mapping at different stages varies. During stage 1 of training process to learn the base network, the rewards of $s$ are used directly. During stage 2 of training process to fine-tune, a static mapping of rewards and actions for $G(s)$ are used; for $t$ the rewards, actions, and states of $t$ itself are used directly.

\begin{figure}[h]
\includegraphics[width= \textwidth]{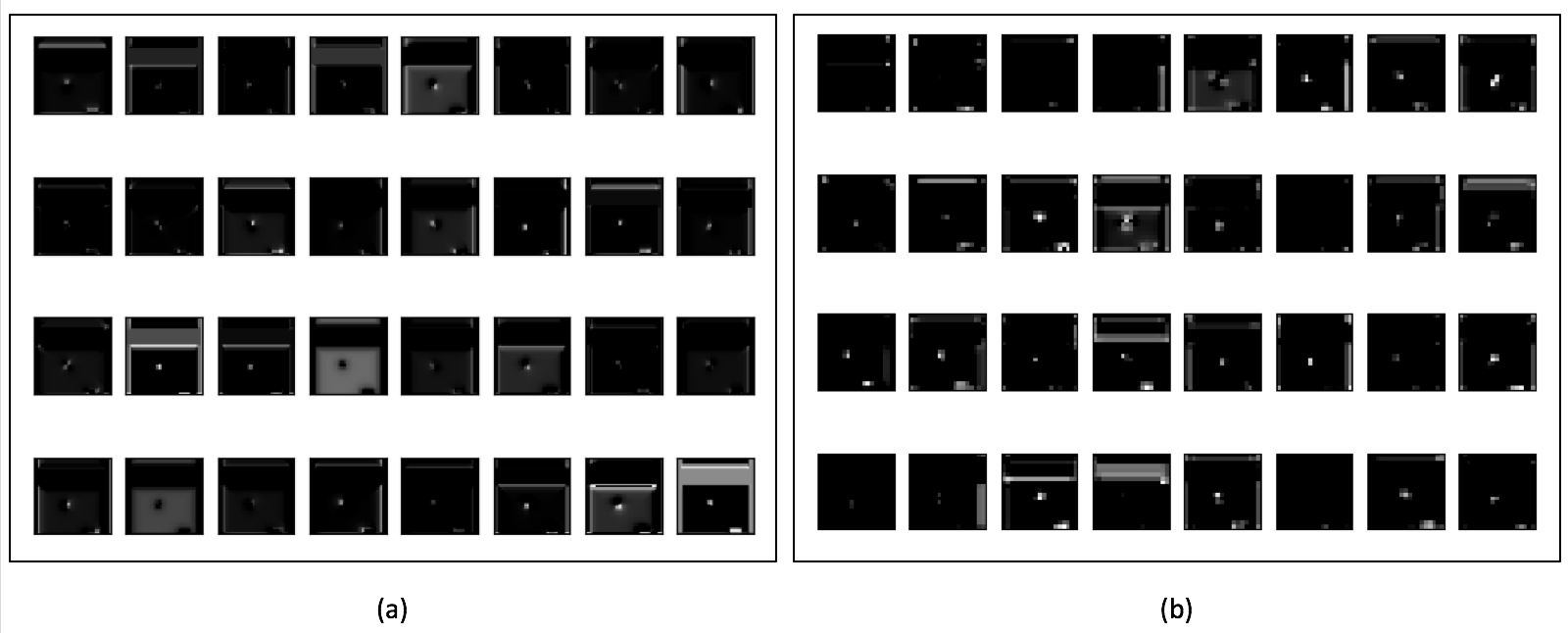}
\caption{(a) Feature activations obtained from the first layer of A3C model trained on \texttt{Breakout-v0} with visual transfer learning method in competitive setting (b) Feature activations obtained from the second layer of A3C model trained on \texttt{Breakout-v0} with visual transfer learning method in competitive setting}
\label{figure5}
\end{figure}

In addition, the activations of states learned by A3C model that is initialized with weights of \texttt{Pong-v0} and fine-tuned for \texttt{Breakout-v0} are shown in Figure \ref{figure5}. The first layer activations are analogous to a series of \textit{Gabor} filters, wherein it learns low level state features like edges, contours in the Atari frames. These include, for instance, the paddle, the position of the fired entity, and the target. The second layer activations extract game specific features by combining low level features from the previously learned layer. In the game play of \texttt{Breakout-v0}, paddle is controlled to hit the ball which determines the rewards and target position does not alter the rewards. This is reflected in the activations of second layer and it can be observed that the pixels at target position are no longer fired. The second layer activations are more focused on information such as the position of the paddle and fired entity; features that are crucial to predict the optimal policies.

\subsection{Transfer Learning using Target Representations}
The first stage of transfer learning process is to train baseline A3C network for source game (\texttt{Pong-v0}) and transfer the knowledge from this model to learning to play target game (\texttt{Breakout-v0}). The source and target games are trained with preprocessed attention frames from Section \ref{gen_inst}. This section compares vanilla transfer learning method of directly using a pre-trained model of source game with the proposed strategy of using competing representations to learn the target game.

\subsubsection{Vanilla Transfer Learning Method }
\begin{figure}[h]
\centering
\begin{minipage}[t]{0.47\textwidth}
\includegraphics[width=\textwidth]{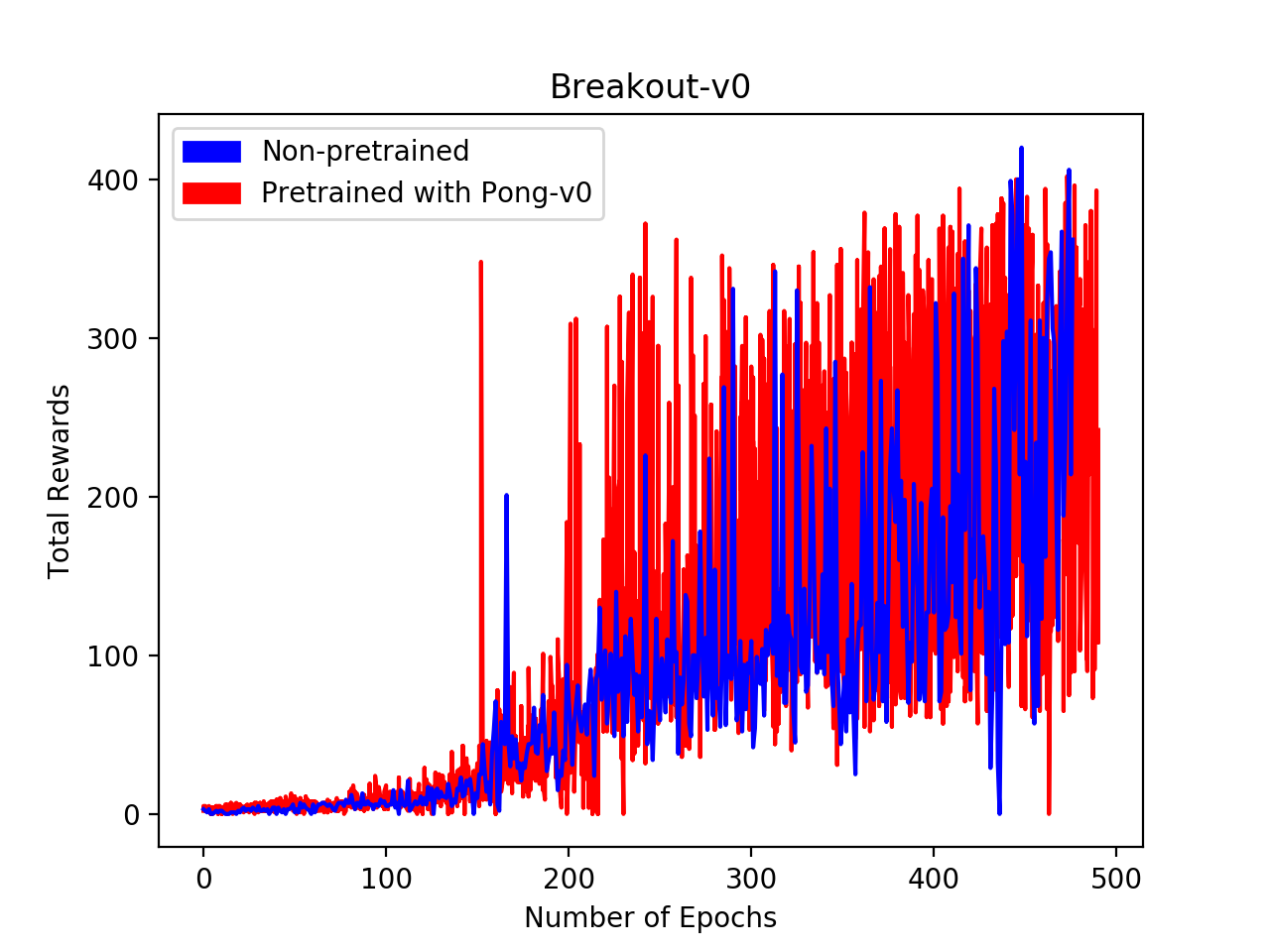}
\end{minipage}
\begin{minipage}[t]{0.47\textwidth}
\includegraphics[width=\textwidth]{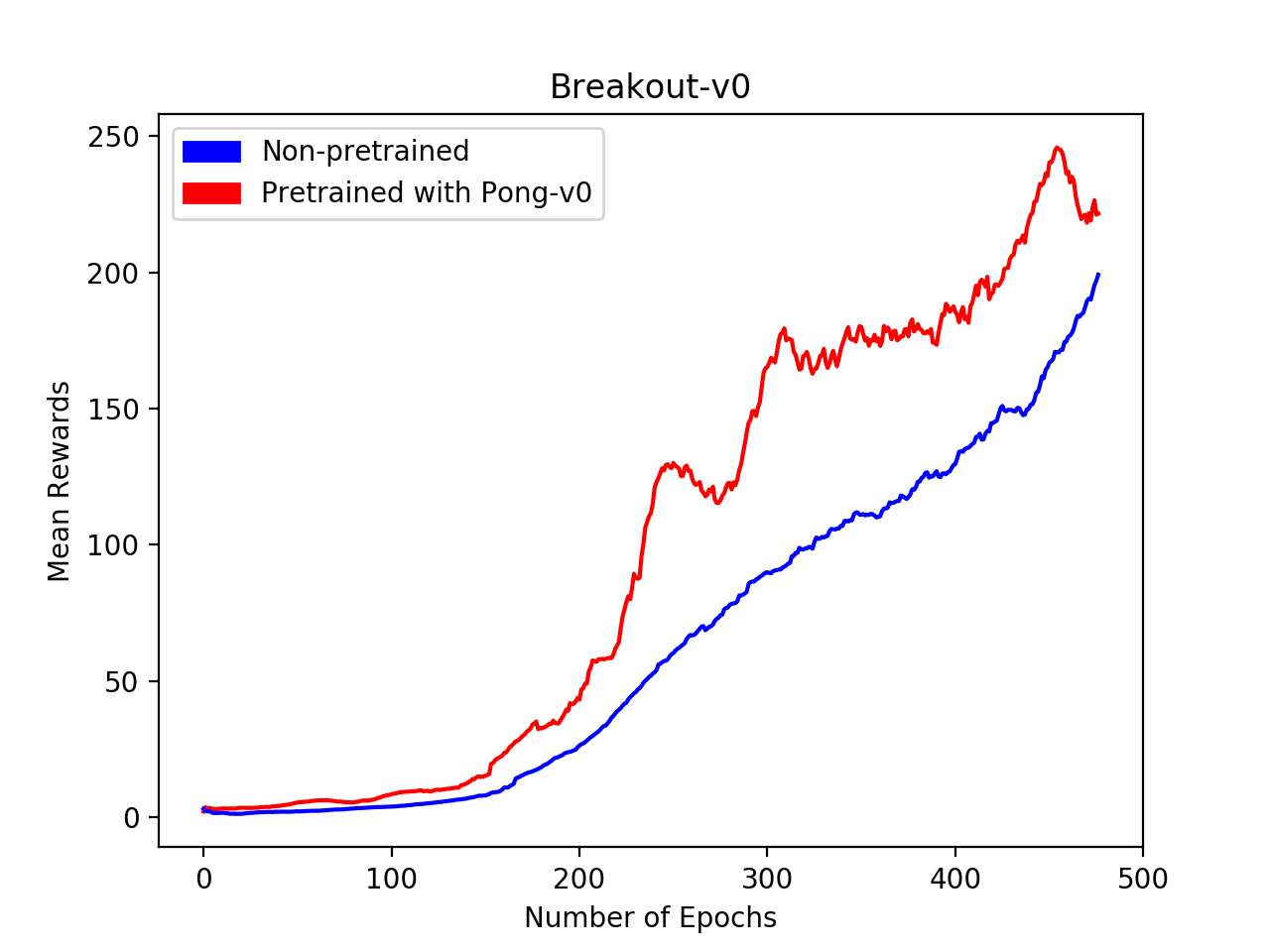}
\end{minipage}
\caption{(a) Total reward per episode versus number of epochs (b) Mean reward per episode over 500 epochs during the training phases of \texttt{Breakout-v0} with and without pre-training.}
\label{figure6}
\end{figure}

The baseline model to be trained for the target game (\texttt{Breakout-v0}) is initialized directly with the weights from the expert model of source game (\texttt{Pong-v0}) excluding the last layer. It is then fine-tuned on the target game to learn the optimal policy. The first graph in Figure \ref{figure6} depicts the training curve of the model fine-tuned with preprocessed frames of \texttt{Breakout-v0}. The blue curve indicates the non pre-trained behavior on \texttt{Breakout-v0}, whereas the red curve indicates the behavior after pre-training. From Figure \ref{figure6}, it is evident that by initializing A3C architecture with weights obtained from \texttt{Pong-v0}, \texttt{Breakout-v0} attains much better rewards as shown in Figure \ref{figure6}(b). Though both the curves reach similar rewards towards the end of training, the  mean rewards obtained by the pre-trained network are significantly higher to show that this transfer learning method works. However, the reward curve rises sharply and fails to make a smooth transition from\texttt{Pong-v0} to \texttt{Breakout-v0} inspite of similar objectives in both the games.    

It can also be seen that the total rewards obtained per episode are not stable and vary a lot between consecutive episodes. With the architecture we propose in Section \ref{gen_inst} using competitive learning between different representations of the target game, we aim to make the model converge faster and smoothen the transition of learning target game from source game. We further evaluate the performance of our models on several evaluation metrics as described in Section \ref{section43}.

\subsubsection{Competitive Transfer Learning Method}
\begin{figure}[h]
\centering
\begin{minipage}[t]{0.47\textwidth}
\includegraphics[width=\textwidth]{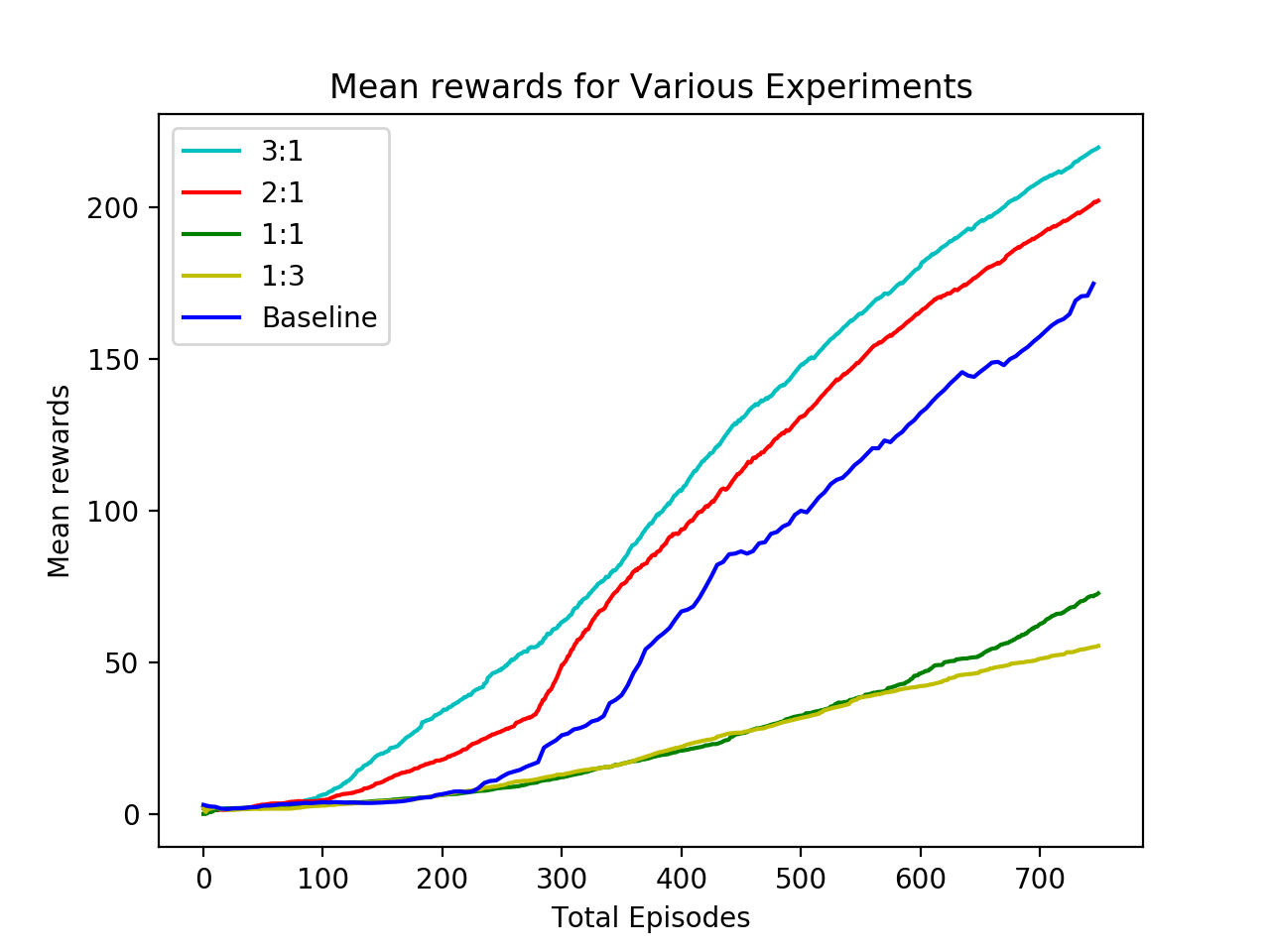}
\end{minipage}
\begin{minipage}[t]{0.47\textwidth}
\includegraphics[width=\textwidth]{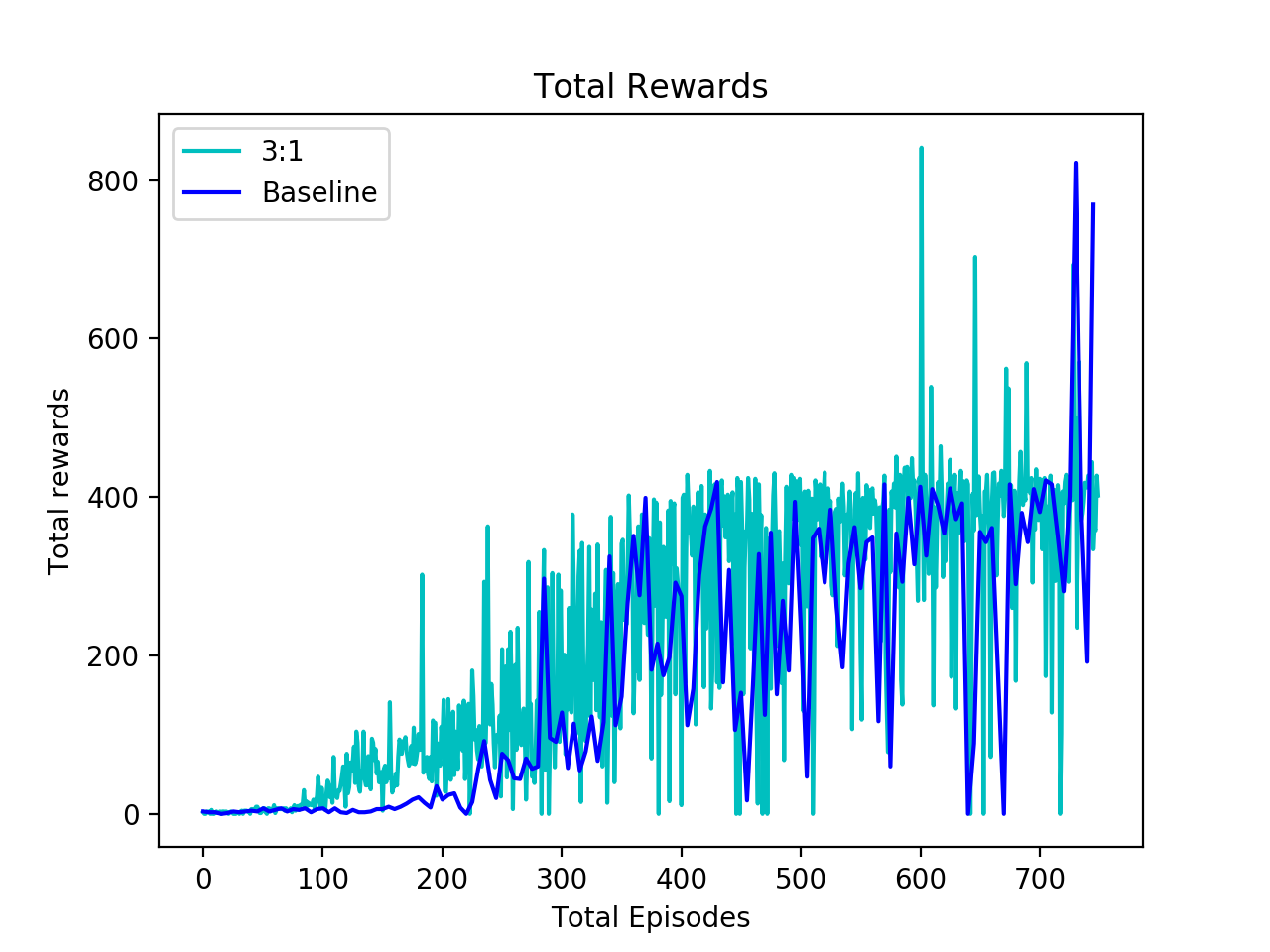}
\end{minipage}
\caption{(a) The mean rewards per episode for the entire set of experiments. The curve in blue is the baseline with no pre-training.(b) Total reward per episode over 700 epochs during the training phases of \texttt{Breakout-v0} with and without pre-training. Here the ratio of workers is 2:1}
\label{figure7}
\end{figure}

In the next stage, we train the agent with a series of representations of the target game that are either queried directly from the environment or generated with the visual mappers from source game as described in Section \ref{gen_inst}. The ratio of number of workers that use these representations simultaneously is a hyper-parameter to be determined through experiments with multiple worker threads. A subset of the workers are fed directly with the frames and actions from target game, \texttt{Breakout-v0} (which are the \textbf{native} frames given by Atari environment). The other subset of workers are fed with frames from the source game, \texttt{Pong-v0} along with the converter (\textbf{visual mapper} to convert source frames to target representations). The results obtained from these experiments are plotted in Figure \ref{figure7}(a). To find the right combination of workers using competing representations, a series of experimentations were carried out with the following ratios of the native vs visual mapper workers as \{3:1, 2:1, 1:1, 1:3\}.

From Figure \ref{figure7}, it is clear that it is possible to learn better representations through the use of pre-trained methods. Further, through experimentations with different configurations of the native versus visual mapper asynchronous workers, it can be concluded that the number of visual mapper workers have a significant impact on the results.

In other experiments, we have also trained a model with the representations derived from the visual mapper (implemented as UNIT GAN convertor) for source game, \texttt{Pong-v0}. This particular model does not get updated with native Atari frames and only uses the generated frames from source game as an input. The model has been trained for over 300 epochs and gained only little improvement in learning the expert policy of target game. Though it did not have significant performance on its own, as it had been established earlier, it acts as a stabilizer to the transfer process with A3C agents.

In the experiments of evaluating running mean rewards plotted in Figure \ref{figure7}(a), the blue curve indicates the baseline model which is trained directly on \texttt{Breakout-v0} environment without using the pre-trained weights obtained from \texttt{Pong-v0}. It is important to note that no converter workers have been used for this setup. The other curves are trained with specific ratios of the native versus visual mapper workers. As can be seen from the Figure \ref{figure7}(a), the configurations with visual mapper workers outnumbering native Atari workers did not perform well. This can be explained due to the inability of the converted workers to learn by themselves as explained previously. On the other hand, as the proportion of these worker decreases, the performance of the model increases. The configurations where the ratio of the workers are 3:1 and 2:1 perform better than the baseline. A detailed discussion on the evaluation of each model will be discussed in Section \ref{section43}.

The improvement of performance of the models with lesser number of visual mapper workers, that use the generated representations of target games, raises questions on its importance. However, upon careful observation of the graph in Figure \ref{figure6}(b), it can be seen that vanilla transfer learning methods, inspite of similarities of both games, do not transition smoothly in learning target game from a certain source game. The aim of these experiments has been to transfer knowledge between games in a stable and data efficient manner. It can also be seen that the final performance of each model with A3C agent converges to the same optimal policy albeit the time taken to converge differs across each experiment setting.

\subsection{Evaluation metrics}\label{section43}
This section evaluates the trained model across a range of metrics \citep{Taylor:2009:TLR:1577069.1755839} for all experiment setups that have been discussed in the previous section.
\begin{table}[h]
\centering
\begin{tabular}{ |p{5cm}|p{1.3cm} |p{1.3cm}|p{1.3cm}| p{1.3cm}|p{1.3cm}| }
 \hline
 \rule{0pt}{3ex}  
 \textbf{Worker Configurations} & \textbf{Original Atari Frames} & \textbf{3:1} & \textbf{2:1} & \textbf{1:1} & \textbf{1:3} \\
 \hline
\rule{0pt}{3ex}  
Jumpstart  &   -  & None & None  & None & None\\
\hline 
\rule{0pt}{3ex}  
Epoch to threshold  &   435  & 357 & 319  & 746 & 872\\
\hline
\rule{0pt}{3ex}  
Total Rewards&  47960   & 74932 & 65376  & 18400 & 17403\\
\hline
\rule{0pt}{3ex}  
Transfer Ratio & - & 1.562 & 1.363 & 0.384 & 0.355 \\
 \hline
\end{tabular}
\caption{Evaluation metrics across different experiment settings with worker configurations (frames taken from native Atari target game vs frames generated from source game using visual mappers)}\label{table1}
\end{table}

The experiments are evaluated with the following metrics to compare the extent of improvement across transfer learning processes. 

\textit{Jumpstart}: This compares the initial performance of an agent in the transfer learning task and it can be seen in Table \ref{table1} that the initial performance could not be improved by transfer from a source task.

\textit{Epoch to threshold}: It measure the time taken to reach a particular level of performance. Based on the results from the graph in Figure \ref{figure7}, it is intuitive to infer that models with higher number of native workers reach a given threshold earlier. Though there isn't a standard threshold level of performance at which the environment is considered to be solved, the threshold for this particular experiment is kept at 400. 

\textit{Total Rewards}: The total rewards are essentially the area under the graph of mean reward per episode vs total number of episodes for each model. It can be inferred, based on the graphs in the previous section, that the agents trained with a higher proportion of native workers have higher total rewards. The mean rewards per episode are only computed for 700 episodes to ensure that the area under the curve is compared across consistent experiment setups.

\textit{Transfer Ratio}: The transfer ratio is the ratio of the total rewards obtained from the transfer learning experiment when compared to the baseline. It measures the effectiveness of transfer learning process. Transfer ratio greater than one implies that the total reward accumulated by the transfer learner is higher than the total reward accumulated by the non-transfer learner and the magnitude specifies the extent of efficient knowledge transfer. Similar to the trends of previous metrics, agents trained with a higher proportion of native workers transfer better. 

\section{Conclusion}
We conclude that it is possible to generate a visual mapper for semantically similar games with the use of UNIT GANs. We then explored the idea of learning two different representations of the same game and using them simultaneously for transfer learning and show that the learning curve is significantly stabilized. Different ratios of the workers were used to study the effect of the visual mapper on transfer learning. Although the workers using representations of the target game obtained from the visual mappers did not perform well in a stand alone setting, however they showed improvements when used for the competitive learning. 

A topic for further research is the generalization of the above discussed methods and techniques to other sets of Atari games. Secondly, the A3C workers that obtain the representations of the source game from the visual mapper; are to be studied and analyzed further to prove the hypothesis that these workers should perform well on both the source and target games simultaneously when allowed to learn models that do well in multiple settings.

The trained models and code can be accessed \href{https://github.com/sowmya-mp/rl_a3c_pytorch}{here for further experiments}.
\medskip

\bibliographystyle{unsrtnat}
\bibliography{references}

\end{document}